%% file: main.tex
\documentclass[runningheads]{llncs}
\usepackage[T1]{fontenc}
\usepackage{graphicx}
\usepackage{booktabs}
\usepackage{multirow}

\usepackage{amsmath}
\usepackage{amssymb}
\usepackage{tikz}
\usetikzlibrary{shapes.geometric}
\usepackage{todonotes}
\usepackage{hyperref}
\usepackage{forest}

\newcommand{\bpmnXOR}{%
  \tikz[baseline=(X.base)]{
    \node[draw, diamond, inner sep=0.2pt, font=\scriptsize] (X) {$\times$};
  }%
}
\newcommand{\bpmnAND}{%
  \tikz[baseline=(X.base)]{
    \node[draw, diamond, inner sep=0.2pt, font=\scriptsize] (X) {$+$};
  }%
}
\newcommand{\bpmnActivity}[1]{%
  \mathord{%
      \begingroup
        \setlength{\fboxsep}{0.6pt}
        \setlength{\fboxrule}{0.1pt}
        \fbox{%
            \makebox[1.8em][c]{\scriptsize$\strut #1$}%
        }%
      \endgroup
  }%
}

\begin{document}
\title{Planning and Scheduling Business Processes under Control-Flow Uncertainty: Extended Version}
\titlerunning{Planning and Scheduling Business Processes under Control-Flow Uncertainty}
%

\author{Michel Kunkler\orcidID{0000-0002-1920-7322},
  Stefanie Rinderle-Ma\orcidID{0000-0001-5656-6108}}

\authorrunning{M. Kunkler, S. Rinderle-Ma}

\institute{Technical University of Munich, Germany \\
  TUM School of Computation, Information, and Technology \\
  \email{\{michel.kunkler,stefanie.rinderle-ma\}@tum.de}
}

%
\maketitle              
\begin{abstract}
  Scheduling activities in business processes can improve efficiency (e.g., reduce makespan), but is challenging because the exact sequence of activities required to complete a case is often uncertain due to decisions based on data that emerges during execution.
  Nevertheless, probabilistic information regarding such decisions can often be estimated or derived from historical execution logs, and can help anticipate which execution paths are likely to lead to successful completion.
  Planning with particular execution paths affects feasibility, i.e., the probability of successful completion, and the expected number of superfluous activities that are planned but never executed.
  We frame the problem as a chance-constrained optimization problem and present two formulations:
  A decomposed approach with two stages, a planning stage that minimizes the expected number of superfluous activities subject to a feasibility constraint, and a scheduling stage that minimizes the makespan over the planned activities; and an integrated approach that combines planning and scheduling into a single formulation.
  Evaluation on two real-world and one synthetic dataset shows that the integrated approach yields superior makespans but is intractable at scale, while the decomposed approach scales to large settings.

\end{abstract}
\input{sections/01_introduction}
\input{sections/02_two_step_approach}
\input{sections/03_integrated_approach}
\input{sections/04_evaluation}
\input{sections/05_related_work}
\input{sections/06_conclusion}

%
%
%
\bibliographystyle{splncs04}
\bibliography{references}
\end{document}

%% file: sections/01_introduction.tex
\section{Introduction}
\label{sec:introduction}
%
%
Scheduling the execution of activities in business processes can improve operational efficiency, e.g., by reducing the makespan of cases and improving resource utilization.
Moreover, schedules provide visibility into the future, which can help resources prepare for their upcoming activities and identify capacity bottlenecks, thereby supporting management decisions regarding business process improvements \cite{DBLP:journals/eor/AytugLMMU05}.
%
Formalizing and solving an optimization problem for scheduling the execution of business process activities is challenging for several reasons: high computational complexity, modeling complexity from control-flow or resource constraints, and the degree of uncertainty in business process execution \cite{DBLP:conf/edoc/KunklerSR25}.
Current work on scheduling activities of business processes has primarily addressed setting up computationally tractable optimization models and capturing a rich set of control-flow and resource constraints \cite{van_der_aalst_petri_1996,DBLP:conf/bpm/SchumannR24}, or addressed uncertain activity durations \cite{DBLP:conf/otm/HavurC19} and resource availability \cite{DBLP:conf/icpm/KunklerR24,DBLP:journals/is/MiddelhuisBSBAD25}.

A key driver of uncertainty in business process execution is that the sequence of activities required to successfully complete a case is often not known with certainty beforehand, e.g., due to runtime constraints or decisions based on data revealed during execution.
Consequently, the exact set of activities that must be scheduled for a case is not known in advance.
While this issue has been acknowledged \cite{DBLP:conf/edoc/KunklerSR25,DBLP:conf/otm/HavurC19} and termed \textsl{control-flow induced uncertainty} \cite{DBLP:conf/edoc/KunklerSR25}, existing work has either focused on scheduling only until the course of the business process can be determined with high certainty \cite{DBLP:conf/otm/HavurC19} or adopted online resource allocation approaches \cite{DBLP:conf/icpm/KunklerR24,DBLP:journals/is/MiddelhuisBSBAD25}.
Both strategies undermine some of the benefits of scheduling, e.g., by preventing resources from preparing for their upcoming activities.

In this work, we address the problem of planning and scheduling the activities of cases under control-flow induced uncertainty until case completion.
We assume that a schedule for a case is feasible if the scheduled activities are sufficient to complete it.
Moreover, we assume that any unnecessarily scheduled activity can be skipped. We refer to such activities as \textsl{superfluous}.
The key idea of this work is to formulate activity selection as a \textsl{chance-constrained optimization problem}, in which activities are selected such that a predefined case feasibility threshold is satisfied while minimizing an optimization objective.
We present two approaches.
In the \textbf{i) decomposed approach}, we first determine the set of activities to be scheduled by solving a planning problem that minimizes the expected number of superfluous activities subject to a feasibility constraint; in the second step, the planned activities are scheduled to minimize the makespan, i.e., the completion time of the final activity.
In the \textbf{ii) integrated approach}, planning and scheduling are combined into a single formulation that directly minimizes the makespan subject to a feasibility constraint.
The decomposed approach allows exclusive activities to overlap, since at most one will actually execute. The integrated approach also allows non-exclusive activities to overlap, which can be beneficial when their joint execution probability is low, keeping the impact on feasibility minor.

\begin{figure}[h]
  \vspace{-5pt}
  \centering
  \includegraphics[width=\linewidth]{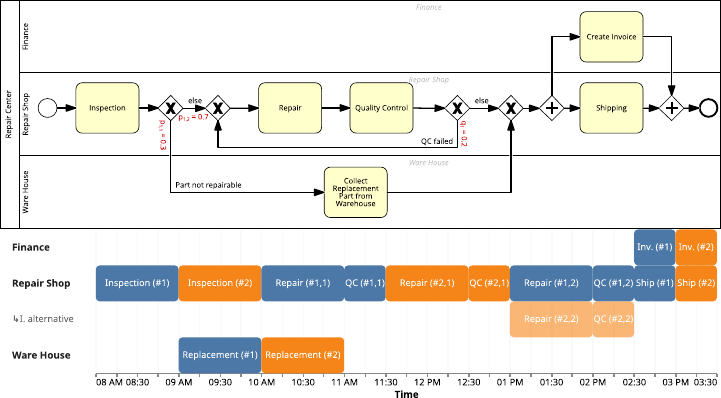} 
  \caption{Exemplary schedule for two cases with overbooking of unlikely activities with a feasibility of $p_{\text{cf}} \approx 0.9295$ and an expected number of feasible cases $\mathbb{E}_{\text{fc}} \approx 1.9283$.}
  \label{fig:example_schedule}
  \vspace{-10pt}
\end{figure}

%
An exemplary business process with control-flow uncertainty is shown in Fig.~\ref{fig:example_schedule}.
It depicts a repair center with an exclusive choice (with two branches and their branching probabilities $p_{1,1}$ and $p_{1,2}$) and a loop (with redo probability $q_{1}$).
Below the process model, an example schedule is shown for two cases and three resources (one per swim lane). In this schedule, the \texttt{Repair} and \texttt{Quality Control (QC)} activities are planned twice for each case.
Since the probability that both cases require a second loop iteration is low, these activities are scheduled to overlap, which can be achieved by the proposed integrated approach.

We formalize both approaches, provide constraint programming implementations and evaluate them on two real-world and one synthetic dataset.
The results show that the integrated approach yields superior makespans but is intractable at scale, while the decomposed approach scales to large settings.

The remainder of this paper is structured as follows.
Sects. \ref{sec:decomposed_approach} and \ref{sec:integrated_approach} present the decomposed and integrated approaches, respectively. Sect. \ref{sec:evaluation} evaluates both approaches, Sect. \ref{sec:related_work} discusses related work, and Sect. \ref{sec:conclusion} concludes.

%% file: sections/02_two_step_approach.tex
\section{Decomposed Planning and Scheduling}
\label{sec:decomposed_approach}

In this section, we formalize the decomposed approach for planning and scheduling activities in business processes under control-flow induced uncertainty. The approach employs stochastic process trees (SPT) as introduced in \cite{DBLP:conf/icpm/CryHB25} due to the decomposition property of the underlying process trees  \cite{DBLP:conf/icsoc/VanhataloVL07} and the augmentation with stochastic information. We reformulate the notation by \cite{DBLP:conf/icpm/CryHB25} because our approach relies on a tree traversal, which requires 1) a unique identifier for each node and 2) a top node.
%

\begin{definition}[Stochastic Process Tree]\label{def:spt}
Let $T$ be a stochastic process tree which consists of sets of operators (sequential $\rightarrow$, parallel $\bpmnAND$, exclusive choices $\bpmnXOR$, redo loops $\circlearrowright$), observable activities $\bpmnActivity{\text{Act}}$, silent transitions $\bpmnActivity{\tau}$ and a designated root $\top$.
Let $\star$ denote the tree's nodes, i.e., $\star := \bpmnAND \cup \bpmnXOR~\cup \circlearrowright \cup \rightarrow \cup~\bpmnActivity{\text{Act}}~\cup~\bpmnActivity{\tau}~\cup \{ \top \}$.
The root $\top$ has exactly one child, $\circlearrowright$ nodes have exactly two ordered children (do-part $ch_1$, redo-part $ch_2$), the other operator nodes ($\rightarrow$, $\bpmnAND$, $\bpmnXOR$) have at least two children and the activity nodes ($\bpmnActivity{\text{Act}}$ and $\bpmnActivity{\tau}$) are leaf nodes.
Let $p_{i,j} \in [0,1]$ denote the transition probability of branch $j$ from an exclusive choice $\bpmnXOR_i$, with $\sum_{j} p_{i,j} = 1$, and $q_i \in [0,1)$ the redo probability of a redo loop $\circlearrowright_i$.
We assume stationarity, i.e., $p_{i,j}$ and $q_i$ remain constant over loop iterations.

We define the following selector functions on a node $\star_i$ of $T$.
Since $T$ is an ordered tree, children of every node are indexed from left to right, starting with 1:
Let $ch(\star_i)$ denote the set of children of $\star_i$;
let $ch_j(\star_i)$ denote the $j$-th child of $\star_i$,
let $par(\star_i)$ denote the parent node of $\star_i$;
let $br(\star_i)$ denote the branch index of $\star_i$ within its parent;
let $anc(\star_i)$ denote the set of strict ancestors of $\star_i$,
let $desc(\star_i)$ denote the descendants of $\star_i$ (including $\star_i$ itself);
and let $ls(\star_i)$ denote the left siblings of $\star_i$.
For two activity nodes $\bpmnActivity{\text{Act}}_i$ and $\bpmnActivity{\text{Act}}_j$, we write $\bpmnActivity{\text{Act}}_i < \bpmnActivity{\text{Act}}_j$ if $\bpmnActivity{\text{Act}}_i$ is to the left of $\bpmnActivity{\text{Act}}_j$ in the process tree, i.e.,
$\bpmnActivity{\text{Act}}_i < \bpmnActivity{\text{Act}}_j \iff \bpmnActivity{\text{Act}}_i \in \bigcup_{k \in anc(\bpmnActivity{\text{Act}}_j) \cup \{\bpmnActivity{\text{Act}}_j\}} \bigcup_{m \in ls(k)} desc(m)$.
\end{definition}
The stochastic process tree for the business process in Fig.~\ref{fig:example_schedule} is shown in Fig.~\ref{fig:running_example_spt}.
\input{figures/running_example_spt}

Given a stochastic process tree, we first solve a chance-constrained planning problem that selects exclusive-choice branches and loop iteration counts subject to a predefined case feasibility threshold.
Because superfluous activities occupy resources at runtime without contributing to the process outcome, we minimize their expected number to reduce unnecessary resource contention in the subsequent scheduling stage.
The planned activities are then scheduled to directly minimize the makespan, exploiting mutual exclusivity to allow overlapping resource assignments.
We develop both stages first for a single case and then extend them to the multi-case setting.

\subsection{The Planning Problem}
Given a stochastic process tree, we define the decision variables for branch selection and loop unrolling, derive the resulting case feasibility and expected number of superfluous activities, and formulate the planning problem as a chance-constrained optimization problem.
\begin{definition}[Decision Variables Planning Problem]
For each exclusive choice node $\bpmnXOR_i$, let $x_{i,j} \in \{0, 1\}$ denote whether branch $j$ is to be planned.
For each loop node $\circlearrowright_i$, let $r_i \in \mathbb{N}^{+}$ denote the number of planned loop iterations, yielding $r_i$ executions of the do-part and $r_i - 1$ executions of the redo-part.
\end{definition}
\begin{definition}[Case Feasibility]
Let T be a stochastic process tree, $x,r$ the decision variables, and $\star_i$ a node in T. Then the node feasibility $p_{\text{nf}}^{(x,r)}(\star_i)$ of $\star_i$ is defined as:
{\small
\begin{equation}
\label{eq:case_feasibility}
\begin{aligned}
p_{\text{nf}}^{(x,r)}(\star_i) &:=
\begin{cases}
p_{\text{nf}}^{(x,r)}(\text{ch}_1(\star_i)),
& \text{if } \star_i = \top, \\
\prod_{j} p_{\text{nf}}^{(x,r)} ( \text{ch}_j (\star_i)),
& \text{if } \star_i \in ~\rightarrow \cup ~\bpmnAND, \\
\sum_{j} x_{i,j}\, p_{i,j}\, p_{\text{nf}}^{(x,r)} ( \text{ch}_j(\star_i) ),
& \text{if } \star_i \in \bpmnXOR, \\
\begin{aligned}[b]
& \bigl( 1 - q_i \bigr) p_{\text{nf}}^{(x,r)}(ch_1(\star_i)) \cdot \\
& \frac{1 - \bigl( q_i p_{\text{nf}}^{(x,r)} ( ch_2(\star_i) ) p_{\text{nf}}^{(x,r)} ( ch_1(\star_i) ) \bigr)^{r_i} }{1 -  q_i p_{\text{nf}}^{(x,r)} ( ch_2(\star_i) ) p_{\text{nf}}^{(x,r)} ( ch_1(\star_i) ) }
\end{aligned},
& \text{if } \star_i \in \circlearrowright, \\
1,
& \text{if } \star_i \in  \bpmnActivity{\tau} \cup \bpmnActivity{\text{Act}}.
\end{cases}
\end{aligned}
\end{equation}
}
Based on the node feasibility $p_{\text{nf}}^{(x,r)}(\star_i)$, we define the case feasibility $p_{\text{cf}}^{(x,r)}$ as:
{\small
\begin{equation}
p_{\text{cf}}^{(x,r)} := p_{\text{nf}}^{(x,r)}(\top)
\end{equation}
}
\end{definition}
We compute $p_{\text{cf}}^{(x,r)}$ top-down from the root $\top$ via the node feasibility $p_{\text{nf}}^{(x,r)}(\star_i)$, the probability that execution of $\star_i$ is feasible given $(x,r)$.
For an exclusive choice $\bpmnXOR_i$, only selected branches contribute, since taking an unselected branch at runtime renders the execution infeasible.
For a loop $\circlearrowright_i$, the do-part executes at least once and each further iteration $k$ contributes with probability $q_i^{k-1}(1-q_i)$, depending on both the do- and redo-parts, yielding
$p_{\text{nf}}^{(x,r)}(\circlearrowright_i) = (1 - q_i)p_{\text{nf}}^{(x,r)}(ch_1(\circlearrowright_i)) + q_ip_{\text{nf}}^{(x,r)}(ch_1(\circlearrowright_i))\sum^{r_i}_{k=2}(q_i^{k-2}(p_{\text{nf}}^{(x,r)}(ch_1(\circlearrowright_i))p_{\text{nf}}^{(x,r)}(ch_2(\circlearrowright_i)))^{k-1}(1-q_i))$,
whose closed form appears in Eq.~\ref{eq:case_feasibility}.

\begin{definition}[Superfluous Activities]\label{def:superfluous} Let T be a stochastic process tree and $x,r$ the decision variables.
The planning count $\nu^{(x,r)}(par(a), a)$ gives the number of times activity $a$ is planned under selection $(x,r)$ (where $par(a)$ denotes the parent of $a$, cf. Def. \ref{def:spt}). 
The expected execution count $\lambda^{(x,r)}(par(a), a)$ gives the expected number of runtime executions of $a$.
The expected number of superfluous activities $s^{(x,r)}$ sums the difference $\nu^{(x,r)} - \lambda^{(x,r)}$ over activities $a \in \bpmnActivity{\text{Act}}$.
{\small
\begin{align}
\label{eq:nu}
\nu^{(x,r)}(\star_i, \star_j) &:=
    \begin{cases}
        1,
        & \text{if } \star_i = \top, \\
        x_{i,br(\star_j)} \nu^{(x,r)}(par(\star_i),\star_i),
        & \text{if } \star_i \in \bpmnXOR, \\
        r_i \nu^{(x,r)}(par(\star_i),\star_i),
        & \text{if } \star_i \in \circlearrowright \land~ br(\star_j) = 1, \\
        (r_i - 1) \nu^{(x,r)}(par(\star_i),\star_i),
        & \text{if } \star_i \in \circlearrowright \land~ br(\star_j) = 2, \\
        \nu^{(x,r)}(par(\star_i),\star_i),
        & \text{else.}
    \end{cases} \\
\label{eq:lambda}
\lambda^{(x,r)}(\star_i, \star_j) &:=
    \begin{cases}
        1,
        & \text{if } \star_i = \top, \\
        x_{i,br(\star_j)} p_{i,br(\star_j)} \lambda^{(x,r)}(par(\star_i),\star_i),
        & \text{if } \star_i \in \bpmnXOR, \\
        \frac{1 - q_{i}^{r_i}}{1 - q_{i}} \lambda^{(x,r)}(par(\star_i),\star_i),
        & \text{if } \star_i \in \circlearrowright \land~ br(\star_j) = 1, \\
        \frac{q_{i} - q_{i}^{r_i}}{1 - q_{i}} \lambda^{(x,r)}(par(\star_i),\star_i),
        & \text{if } \star_i \in \circlearrowright \land~ br(\star_j) = 2, \\
        \lambda^{(x,r)}(par(\star_i),\star_i),
        & \text{else.}
    \end{cases} \\
\label{eq:superfluous}
s^{(x,r)} &:= \sum_{a \in \bpmnActivity{\text{Act}}}\bigl(\nu^{(x,r)}(par(a), a) - \lambda^{(x,r)}(par(a), a)\bigr)
\end{align}
}
\end{definition}
We compute $\nu^{(x,r)}$ and $\lambda^{(x,r)}$ bottom-up; both depend on the activity's exclusive choice and loop ancestors.
For $\nu^{(x,r)}$, the activity counts as planned only if every $\bpmnXOR_i$ ancestor is selected. Its base value $1$ is multiplied by $r_i$ for each loop ancestor in whose do-part it lies, and by $r_i-1$ for each in whose redo-part it lies.
For $\lambda^{(x,r)}$, the base value $1$ is multiplied by $p_{i,j}x_{i,j}$ at each $\bpmnXOR_i$ ancestor and by the expected iteration count $\sum_{k=1}^{r_i} q_i^{k-1}$ (do-part) or $\sum_{k=2}^{r_i} q_i^{k-1}$ (redo-part) for each loop ancestor; both sums admit closed forms used in Eq.~\ref{eq:lambda}.

Fig.~\ref{fig:running_example_feasibility} illustrates the node feasibilities for the running example as well as the planned and expected execution counts. 

\input{figures/running_example_feasibility}
\subsubsection{Solving the Planning Problem}
Based on the above, we can write the planning problem as a mixed-integer nonlinear program (MINLP) (Eq. \ref{eq:minlp}) that minimizes the expected number of superfluous activities while satisfying a case feasibility threshold $\theta \in (0, 1]$.

\begin{equation}
\label{eq:minlp}
\vspace{-5pt}
\begin{aligned}
\min_{x,r}\quad & s^{(x,r)} \\
\text{s.t.}\quad & p_{\text{cf}}^{(x,r)} \ge \theta \\
                 & x_{i,j} \in \{0, 1\} && \forall \bpmnXOR_i, \forall j \\
                 & r_i \in \mathbb{N}^{+} && \forall \circlearrowright_i \\
\end{aligned}
\end{equation}

\subsection{The Scheduling Problem}
Given a selection $(x, r)$ from the planning problem, we construct a scheduling problem that minimizes the makespan of the planned activities, i.e., the time until the end of the last activity in a case, e.g., makespan $=7$ for the blue case in Fig. \ref{fig:example_schedule}.
To this end, we unroll the process tree and remove non-selected branches to obtain the set of activities and their precedence constraints and their mutual exclusivity.
We then identify activities that are mutually exclusive; since at most one of them will be executed at runtime, they can be assigned to the same resource concurrently.
Finally, we formulate the scheduling problem as a MINLP.
\subsubsection{Unrolled and Branch Selected Process Tree}
\label{sec:sub:unrolling_and_selection}
Given a selection $(x, r)$, we transform the process tree by removing non-selected branches and unrolling loops.
For every $x_{i,j} = 0$, the node $ch_j(\bpmnXOR_i)$ and its descendants are removed.
Each loop node $\circlearrowright_i$ is unrolled according to $r_i$: if $r_i = 1$, it is replaced by its do-part $ch_1(\circlearrowright_i)$.
For $r_i > 1$, it is replaced by a sequential composition beginning with the do-part, where each subsequent iteration is an exclusive choice between a silent transition $\bpmnActivity{\tau}$ (loop exit) and the redo-part followed by the do-part and the same choice recursively, terminating after $r_i$ iterations.
Any schedule must respect the precedence orders set out by the original process tree; this equally applies to the unrolled and branch-selected process tree. Otherwise, important functional or compliance ordering constraints would be violated, e.g., examining the patient before surgery. Hence, an activity $\star_i$ can only start after all its predecessors $pr(\star_i)$ have completed where
$pr(\star_i)=desc(\star_j) \cap \bpmnActivity{\text{Act}}$, i.e., all descendants of all left siblings $\star_j \in ls(\star_k)$, where $\star_k$ is $\star_i$ itself or any ancestor of $\star_i$ with $par(\star_k) \in \rightarrow$.

Since at most one branch of an exclusive choice is executed at runtime, mutually exclusive activities can share a resource in overlapping time slots.

\begin{definition}[Mutual Exclusivity]\label{eq:mutual_exclusivity}
Two activities $a_i, a_j$ are mutually exclusive ($\bpmnXOR(a_i, a_j) = 1$) if there exists a common exclusive-choice ancestor $\bpmnXOR_k \in anc(a_i) \cap anc(a_j) \cap \bpmnXOR$ with distinct children $m \neq n$ such that $a_i \in desc(ch_m(\bpmnXOR_k))$ and $a_j \in desc(ch_n(\bpmnXOR_k))$. The set of activities exclusive to $a_i$ is $\bpmnXOR(a_i) := \{a_j \in \bpmnActivity{\text{Act}} \mid \bpmnXOR(a_i, a_j) = 1\}$.
\end{definition}
\subsubsection{Scheduling Formulation}
We formalize the scheduling problem as a joint resource-assignment and activity-start-time decision problem to minimize the makespan.
For each activity $a \in \bpmnActivity{\text{Act}}$, let $I_a$ be an interval variable with accessors $\text{start}(I_a)$, $\text{end}(I_a)$, $\text{dur}(I_a)$, and $\text{overlapping}(I_a, I_{a'}) \in \{0,1\}$ which equals $1$ iff $I_a$ and $I_{a'}$ share a common time point.
Let $\mathcal{R}$ be the set of resources, and for each $(a,\rho)\in \bpmnActivity{\text{Act}}\times \mathcal{R}$, let $d_{a,\rho}$ be the processing time of $a$ on $\rho$ and $z_{a,\rho} \in \{0, 1\}$ the allocation variable.
The formulation in Eq.~\ref{eq:scheduling} minimizes the makespan~\eqref{eq:scheduling:objective}, sets the processing time per resource assignment~\eqref{eq:scheduling:dur}, enforces precedence~\eqref{eq:scheduling:prec}, ensures each activity is assigned to exactly one resource~\eqref{eq:scheduling:act}, and prevents temporal overlap of non-mutually-exclusive activities on the same resource~\eqref{eq:scheduling:cap}.
{\small
\begin{subequations}\label{eq:scheduling}
\begin{align}
\min_{z,I}\quad  & \max (\left\{\text{end}(I_a)\mid a \in \bpmnActivity{\text{Act}} \right\})
\label{eq:scheduling:objective}
\\
\text{s.t.}\quad &
\text{dur}(I_a) = \sum_{\rho \in \mathcal{R}}z_{a,\rho}d_{a,\rho}
&& \forall a \in \bpmnActivity{\text{Act}}
\label{eq:scheduling:dur}
\\
&
\text{start}(I_a) \geq \text{end}(I_{a'})
&& \forall a \in \bpmnActivity{\text{Act}}, \forall a' \in pr(a)
\label{eq:scheduling:prec}
\\
&
\sum_{\rho \in \mathcal{R}}z_{a,\rho} = 1
&& \forall a \in \bpmnActivity{\text{Act}}
\label{eq:scheduling:act}
\\
&
z_{a,\rho} \cdot z_{a',\rho} \cdot \text{overlapping}(I_a, I_{a'}) = 0
&&
\begin{aligned}[t]
  & \forall a, a' \in \bpmnActivity{\text{Act}} : a \neq a', \\
  & \quad \bpmnXOR(a, a') = 0, \;\; \forall \rho \in \mathcal{R}
\end{aligned}
\label{eq:scheduling:cap}
\end{align}
\end{subequations}
}
\subsection{Multi-Case Planning and Scheduling}
\begin{figure}[h]
    \vspace{-10pt}
    \centering
    \includegraphics[width=0.8\textwidth]{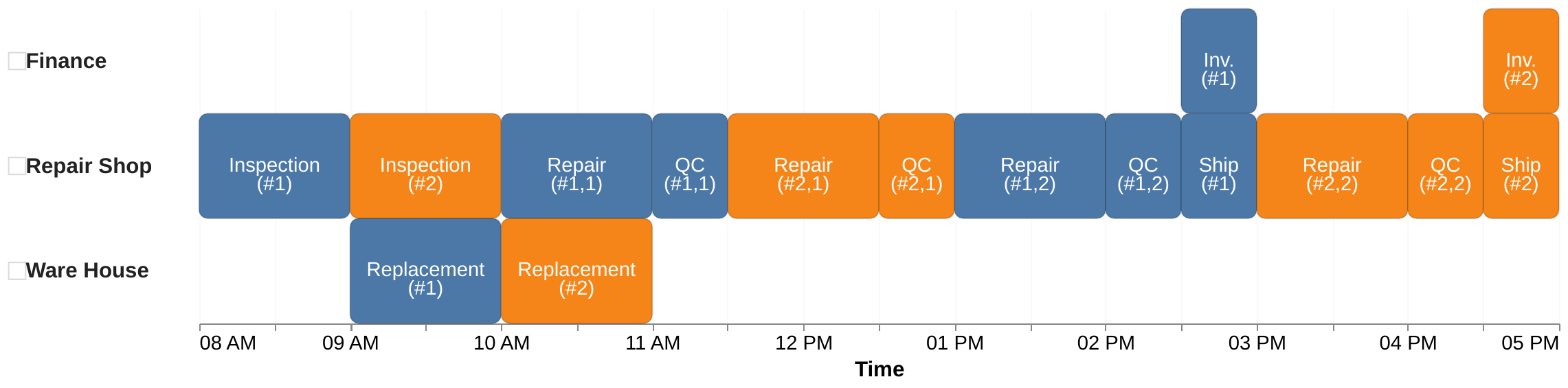}
    \caption{Schedule generated by the decomposed approach for the running example with two cases.}
    \label{fig:decomposed_schedule}
    \vspace{-15pt}
\end{figure}

In practice, multiple cases must be scheduled concurrently, so we extend the decomposed approach to $n$ cases.
The per-case feasibility $p_{\text{cf}}$ does not directly convey how many cases are expected to complete successfully; we therefore adopt the expected number of feasible cases $\mathbb{E}_{\text{fc}}$ as the chance constraint.
We reuse the MINLP (Eq.~\ref{eq:minlp}) with per-case threshold $\theta = E/n$, where $E$ denotes the required expected number of feasible cases, and apply the resulting selection $(x,r)$ uniformly to all $n$ cases.
Since precedence and mutual exclusivity apply within each case, the scheduling formulation extends to $n$ cases by taking the union of the $n$ unrolled process trees.
A schedule from the decomposed approach for the two cases from the running example is shown in Fig.~\ref{fig:decomposed_schedule}.
A lower number of superfluous activities could in principle be achieved by per-case planning, i.e., composing $n$ copies of the process tree in parallel, but this renders the planning problem intractable for large $n$; the integrated approach in Sect.~\ref{sec:integrated_approach} addresses this by jointly optimizing planning and scheduling.

%% file: figures/running_example_spt.tex
\begin{figure}[h]
\vspace{-10pt}
\centering
\begin{forest}
  for tree={
    align=center,
    l sep=0.8mm,
    l=0pt,
    s sep=19mm,
    inner sep=1.8pt,
    edge={-},
  },
  gate/.style={
    draw, diamond, aspect=1.2,
    inner sep=1pt,
    font=\scriptsize,
    minimum size=1.4em,
  },
  act/.style={
    draw, rectangle,
    font=\scriptsize,
    inner sep=1pt,
    minimum height=1.3em,
    minimum width=1.8em,
  },
  oper/.style={font=\small},
  [$\top$, oper
    [$\rightarrow$, oper
      [{insp.}, act]
      [$\times$, gate
        [$\circlearrowright$, oper,
          edge label={node[midway, left, font=\scriptsize]{$p_{1,1}{=}0.7$}}
          [$\rightarrow$, oper
            [{rep.}, act]
            [{q.c.}, act]
          ]
          [$\tau$, act,
            edge label={node[midway, right, font=\scriptsize]{$q_1{=}0.2$}}
          ]
        ]
        [{col.}, act,
          edge label={node[midway, right, font=\scriptsize]{$p_{1,2}{=}0.3$}}
        ]
      ]
      [$+$, gate
        [{ship.}, act]
        [{inv.}, act]
      ]
    ]
  ]
\end{forest}
\caption{Stochastic process tree for the repair process.}
\label{fig:running_example_spt}
\vspace{-10pt}
\end{figure}

%% file: figures/running_example_feasibility.tex
\begin{figure}[h]
\vspace{-15pt}
\centering
\begin{forest}
  for tree={
    align=center,
    l sep=0.8mm,
    l=0pt,
    s sep=19mm,
    inner sep=1.8pt,
    edge={-},
  },
  gate/.style={
    draw, diamond, aspect=1.2,
    inner sep=1pt,
    font=\scriptsize,
    minimum size=1.4em,
  },
  act/.style={
    draw, rectangle,
    font=\scriptsize,
    inner sep=1pt,
    minimum height=1.3em,
    minimum width=1.8em,
  },
  oper/.style={font=\small},
  [$\top$, oper, label={[font=\scriptsize, text=red, inner sep=0.5pt, label distance=0.5pt]above:$0.972$}
    [$\rightarrow$, oper, label={[font=\scriptsize, text=red, inner sep=0.5pt, label distance=0.5pt]above:$0.972$}
      [{insp.}, act, label={[font=\scriptsize, text=blue, inner sep=0.5pt, label distance=0.5pt]below:$\nu{=}1,\;\lambda{=}1$}]
      [$\times$, gate, label={[font=\scriptsize, text=red, inner sep=0.5pt, label distance=5pt]above:$0.972$}
        [$\circlearrowright$, oper,
          edge label={node[midway, left, font=\scriptsize]{$x_{1,1}{=}1$}},
          label={[font=\scriptsize, text=red, inner sep=0.5pt, label distance=0.5pt]left:$0.96$},
          label={[font=\scriptsize, inner sep=0.5pt, label distance=0.5pt]right:$r_1{=}2$}
          [$\rightarrow$, oper, label={[font=\scriptsize, text=red, inner sep=0.5pt, label distance=0.5pt]above:$1$}
            [{rep.}, act, label={[font=\scriptsize, text=blue, inner sep=0.5pt, label distance=0.5pt]below:$\nu{=}2,\;\lambda{=}0.84$}]
            [{q.c.}, act, label={[font=\scriptsize, text=blue, inner sep=0.5pt, label distance=0.5pt]below:$\nu{=}2,\;\lambda{=}0.84$}]
          ]
          [$\tau$, act, label={[font=\scriptsize, text=blue, inner sep=0.5pt, label distance=0.5pt]below:$\nu{=}1,\;\lambda{=}0.14$}]
        ]
        [{col.}, act,
          edge label={node[midway, right, font=\scriptsize]{$x_{1,2}{=}1$}},
          label={[font=\scriptsize, text=blue, inner sep=0.5pt, label distance=0.5pt]below:$\nu{=}1,\;\lambda{=}0.3$}
        ]
      ]
      [$+$, gate, label={[font=\scriptsize, text=red, inner sep=0.5pt, label distance=0.5pt]above:$1$}
        [{ship.}, act, label={[font=\scriptsize, text=blue, inner sep=0.5pt, label distance=0.5pt]below:$\nu{=}1,\;\lambda{=}1$}]
        [{inv.}, act, label={[font=\scriptsize, text=blue, inner sep=0.5pt, label distance=0.5pt]below:$\nu{=}1,\;\lambda{=}1$}]
      ]
    ]
  ]
\end{forest}
\caption{Stochastic process tree with selection $(x,r)$, node feasibilities in \textcolor{red}{red}, planned ($\nu$) and expected executed ($\lambda$) counts in \textcolor{blue}{blue}.}
\label{fig:running_example_feasibility}
\label{fig:running_example_superfluous}
\vspace{-30pt}
\end{figure}

%% file: sections/03_integrated_approach.tex
\section{Integrated Approach}
\label{sec:integrated_approach}
The decomposed approach has two limitations that can affect the optimality of the resulting schedule, which we address in the integrated approach:
First, minimizing superfluous activities in the planning stage does not necessarily lead to a minimal makespan in the scheduling stage, as selecting a branch with more (superfluous) activities can be beneficial for the makespan, e.g., when this avoids scheduling on a bottleneck resource.
Second, the decomposed approach only allows exclusive activities to be scheduled concurrently on the same resource.
Additionally allowing non-exclusive activities to overlap on a resource can further shorten the makespan.
This holds especially in the multi-case setting where it can be meaningful to, e.g., schedule two activities from two different cases that both have low execution probability concurrently as this will only lead to a minor decrease of $\mathbb{E}_{\text{fc}}$ (see Fig.~\ref{fig:example_schedule}).
We therefore present an integrated approach that combines planning and scheduling into a single formulation, allowing per-case selections and the overlapping scheduling of non-exclusive activities.

\begin{definition}[Decision Variables Integrated Problem]
For each case $\kappa \in K$ and each exclusive choice node $\bpmnXOR_i$, let $x^{\kappa}_{i,j} \in \{0,1\}$ indicate whether branch $j$ is selected for $\kappa$.
For each loop node $\circlearrowright_m$, let $r^{\kappa}_m \in \mathbb{N}^+$ denote the number of planned iterations, yielding $r^{\kappa}_m$  executions of the do-part and $r^{\kappa}_m - 1$ executions of the redo-part.
\end{definition}

We obtain unrolled-and-selected process trees, from which we directly derive precedence and mutual exclusivity relationships and compute the set of activities for a scenario and its probability.

\begin{definition}[Unrolled-and-Selected Process Tree]\label{def:unrolled_selected_tree}
For case $\kappa$ and selection $(x^{\kappa}, r^{\kappa})$, the unrolled-and-selected process tree
$T^{(x^{\kappa}, r^{\kappa})}$
is obtained from $T$ by:
(i) deleting, for every $\bpmnXOR_i$ and every branch $j$ with
$x^{\kappa}_{i,j} = 0$ the subtree rooted at $ch_j(\bpmnXOR_i)$, while retaining $\bpmnXOR_i$ itself even if only one branch remains;
(ii) replacing the do-subtree of each loop $\circlearrowright_m$ by a sequential composition of
$r^{\kappa}_m$ unrolled do-copies interleaved with
$r^{\kappa}_{m} - 1$ unrolled redo-copies (do, redo, do, $\dots$, redo, do), and its redo-subtree by a placeholder $\bpmnActivity{\tau}$ that is never visited at runtime.
We write $\hat{v}$ for nodes of
$T^{(x^{\kappa}, r^{\kappa})}$ and $t(\hat{v})$ for the original-tree node that $\hat{v}$ instantiates (undefined on the placeholder $\tau$).
\end{definition}
Unlike the unrolling in Sect.~\ref{sec:decomposed_approach}, we keep every $\bpmnXOR_i$, flatten the loop, and retain $\circlearrowright_m$ rather than introducing exit-$\bpmnXOR$s, so that every $\bpmnXOR$ and $\circlearrowright$ in
$T^{(x^{\kappa}, r^{\kappa})}$ carries the original $p_{i,j}$ or $q_m$ and the runtime draws at these nodes can be evaluated per case (Def.~\ref{def:execution_scenario}).

\begin{definition}[Planned Activities]\label{def:planned_activities}
The set of planned activities under selection
$(x^{\kappa}, r^{\kappa})$ is
\begin{equation}
A^{(x^{\kappa}, r^{\kappa})} := \{a_{\kappa, \hat{v}} \mid \hat{v} \in T^{(x^{\kappa}, r^{\kappa})},\, t(\hat{v}) \in \bpmnActivity{\text{Act}}\}, \quad A^{(x,r)} := \bigcup_{\kappa \in K} A^{(x^{\kappa}, r^{\kappa})},
\end{equation}
with $t(a_{\kappa, \hat{v}}) := t(\hat{v})$.
For convenience, we write $a_\alpha$ for $a_{\kappa, \hat{v}}$ with $\alpha = (\kappa, \hat{v})$.
We lift the precedence constraint in Eq.~\ref{eq:scheduling:prec} to planned activities by $pr(a_{\kappa, \hat{v}}) := \{a_{\kappa, \hat{v}'} \mid \hat{v}' \in pr(\hat{v})\}$, with $pr$ applied to $T^{(x^{\kappa}, r^{\kappa})}$.
\end{definition}

\subsubsection{Expected Case Feasibility with Overlapping Activities}
Since the integrated formulation permits non-exclusive activities to overlap on the same resource, case feasibility no longer depends solely on branch selection and loop unrolling: for non-exclusive overlapping activities, multiple activities may need to execute, reducing $\mathbb{E}_{\text{fc}}$.
We formalize this by defining (i) an activity priority order that determines which activity runs when overlapping allocations conflict, (ii) the execution probabilities of overlapping activities, and (iii) a lower bound on the expected number of feasible cases.

\begin{definition}[Activity Priority Order]
When two planned activities $a_\alpha = a_{\kappa, \hat{v}}$ and $a_{\alpha'} = a_{\kappa', \hat{v}'}$ are allocated to overlap on the same resource, only the higher-priority one can execute; we say $a_\alpha$ has higher priority than $a_{\alpha'}$ iff $a_\alpha < a_{\alpha'}$, with:
\begin{equation}
\begin{aligned}
a_{\alpha} < a_{\alpha'} \iff & \Bigl( \kappa < \kappa' \Bigr) \;\lor\; \Bigl( \kappa = \kappa' \land  a_{\alpha} \in pr(a_{\alpha'}) \Bigr) \;\lor\; \\
& \Bigl( \kappa = \kappa' \land a_{\alpha'} \notin pr(a_{\alpha}) \land a_{\alpha} \notin pr(a_{\alpha'}) \land \hat{v} < \hat{v}'  \Bigr)
\end{aligned}
\end{equation}
\end{definition}

\begin{definition}[Overlapping Activities]
Two activities $a_{\alpha}, a_{\alpha'}$ are overlapping if they are allocated to the same resource and have overlapping time intervals.
The set of activities overlapping with $a_\alpha$ given intervals $I$ and resource assignments $z$ is denoted $O^{(x,r,I,z)}(a_\alpha)$.
\begin{equation}
\begin{aligned}
O^{(x,r,I,z)}(a_{\alpha}) := \left\{a_{\alpha'} \in A^{(x,r)} \setminus \{a_{\alpha}\} \;\middle|\;
    \begin{aligned}
        & \text{overlapping}(I_{a_{\alpha}}, I_{a_{\alpha'}}) \land \\
        & \sum_{\rho \in \mathcal{R}}(z_{a_\alpha,\rho}z_{a_{\alpha'},\rho}) = 1
    \end{aligned}
\right\}
\end{aligned}
\end{equation}
\end{definition}
Since an activity can overlap with multiple activities from another case, and those activities may have correlated executions, we can no longer treat each overlap as an independent threat to case feasibility.
To capture the joint execution probabilities of overlapping activities, we introduce a per-case scenario-based approach which exploits three structural facts: (i) scenarios across cases are independent; (ii) exclusive planned activities can never be executed together; (iii) within each scenario, activity executions are deterministic, resolving any remaining intra-case dependencies.
\begin{definition}[Execution Scenario]\label{def:execution_scenario}
An execution scenario $\omega$ for case $\kappa \in K$ is a runtime realization drawing, independently at each reached visit, a branch $b_{\hat{i}}^\omega \in ch(t(\hat{i}))$ of the original tree at every $\hat{i}$ with $t(\hat{i}) \in \bpmnXOR$ and an iteration count $n_{\hat{m}}^\omega \in \mathbb{N}^+$ at every $\hat{m}$ with $t(\hat{m}) \in \circlearrowright$; let $\hat{\mathcal{X}}(\omega)$ and $\hat{\mathcal{L}}(\omega)$ denote the sets of $\bpmnXOR$ and $\circlearrowright$ nodes visited in $\omega$.
$\omega$ is planning-feasible, $\omega \in \Omega_\kappa^{(x,r)}$, iff all draws fit the plan: $x^{\kappa}_{t(\hat{i}), b_{\hat{i}}^\omega}=1$ and $n_{\hat{m}}^\omega \le r^{\kappa}_{t(\hat{m})}$.
$\omega$ is plan-bounded, $\omega \in \bar{\Omega}_\kappa^{(x,r)}$, iff only its loop draws fit the plan, $n_{\hat{m}}^\omega \le r^{\kappa}_{t(\hat{m})}$, while any branch of $t(\hat{i})$ may be drawn: if $x^{\kappa}_{t(\hat{i}), b_{\hat{i}}^\omega}=0$, the drawn subtree is absent from $T^{(x^{\kappa}, r^{\kappa})}$, and the traversal ignores this plan failure, i.e., it visits no node below $\hat{i}$ and continues after $\hat{i}$.
Both sets are finite despite the infinite scenario space, and $\Omega_\kappa^{(x,r)} \subseteq \bar{\Omega}_\kappa^{(x,r)}$.
For $\omega \in \bar{\Omega}_\kappa^{(x,r)}$, its probability and visited planned activities are:
\begin{equation}
\begin{aligned}
p_s(\omega) &:= \prod_{\hat{i} \in \hat{\mathcal{X}}(\omega)} p_{t(\hat{i}),\, b_{\hat{i}}^\omega} \cdot \prod_{\hat{m} \in \hat{\mathcal{L}}(\omega)} q_{t(\hat{m})}^{n_{\hat{m}}^\omega - 1}(1 - q_{t(\hat{m})}) \\
A_\kappa^{(x,r)}(\omega) &:= \{a_{\kappa, \hat{v}} \in A_\kappa^{(x,r)} \mid \hat{v} \text{ visited in } \omega\}
\end{aligned}
\end{equation}
\end{definition}
\begin{definition}[Exclusive Planned Activities]\label{def:exclusive}
We lift the mutual exclusivity relation of Eq.~\ref{eq:mutual_exclusivity} (applied to $T^{(x^{\kappa}, r^{\kappa})}$) to planned activities; two are exclusive only if they belong to the same case:
\begin{equation}
\bpmnXOR(a_{\kappa, \hat{v}}, a_{\kappa', \hat{v}'}) := \mathbf{1}[\kappa = \kappa'] \cdot \bpmnXOR(\hat{v}, \hat{v}'), \, \bpmnXOR(a_{\alpha}) := \{ a_{\alpha'} \mid \bpmnXOR(a_{\alpha}, a_{\alpha'}) = 1 \}.
\end{equation}
\end{definition}
\begin{definition}[Threats and Intra-Case Feasibility]\label{def:threats_chi}
For an activity set $A' \subseteq A^{(x,r)}$, let $\mathcal{T}(A') \subseteq A^{(x,r)}$ contain the threats to activities in $A'$, i.e., activities that overlap with some $a \in A'$, have higher priority than $a$, and are non-exclusive to $a$.
Further, let the intra-case feasibility indicator $\chi^{(x,r,I,z)}(\kappa, \omega)$ be $1$ iff scenario $\omega$ of case $\kappa$ contains no internal conflict, i.e., no activity of $\kappa$ is threatened by another non-exclusive activity of $\kappa$. Formally:
\begin{equation}
\label{eq:threats_and_chi}
\begin{aligned}
    \mathcal{T}^{(x,r,I,z)}(A') &:= \bigcup_{a \in A'} \bigl(\{ f \in O^{(x,r,I,z)}(a) \mid f < a \} \setminus \bpmnXOR(a)\bigr) \\
    \chi^{(x,r,I,z)}(\kappa, \omega) &:= \mathbf{1}\!\left[\mathcal{T}^{(x,r,I,z)}(A_\kappa^{(x,r)}(\omega)) \cap A_\kappa^{(x,r)}(\omega) = \emptyset\right]
\end{aligned}
\end{equation}
\end{definition}

Building on the above definitions, we can now compute the expected number of feasible cases.

\begin{definition}[Expected Feasible Cases]
For a case $\kappa$ with threat activities $C$, the case threat probability $p_{\text{ctp}}^{(x,r)}\allowbreak(\kappa', C)$ is the probability of a potential threat from another case $\kappa'$, i.e., that $\kappa'$ executes at least one of its own activities in $C$ before its own infeasibility, if any, becomes apparent; it sums over the plan-bounded scenarios $\bar{\Omega}_{\kappa'}^{(x,r)}$ with a tail-absorbing scenario weight $\tilde{p}_s$ that differs from $p_s$ only at the maximum loop iteration.
The overlap-adjusted case feasibility $p_{\text{cfo}}(\kappa)$ is the probability that case $\kappa$ completes without a realized threat; it sums $p_s(\omega)$ over planning-feasible scenarios, with $\chi(\kappa, \omega)$ handling intra-case threats and a product over $\kappa' \neq \kappa$ handling cross-case threats.
The expected number of feasible cases $\mathbb{E}_{\text{fc}}$ sums $p_{\text{cfo}}$ over cases; it is a lower bound, since a cross-case higher-priority activity need not be executed when its case is already infeasible, and since potential threats are counted even after their own case has failed.
{\small
\begin{equation}
\label{eq:expected_feasible_cases}
\begin{aligned}
    \tilde{p}_s(\omega) &:= \prod_{\hat{i} \in \hat{\mathcal{X}}(\omega)} p_{t(\hat{i}),\, b_{\hat{i}}^\omega} \cdot \prod_{\hat{m} \in \hat{\mathcal{L}}(\omega)} q_{t(\hat{m})}^{\,n_{\hat{m}}^\omega - 1}\,(1 - q_{t(\hat{m})})^{\mathbf{1}[n_{\hat{m}}^\omega < r^{\kappa'}_{t(\hat{m})}]} \\
    p_{\text{ctp}}^{(x,r)}(\kappa', C) &:= \sum_{\omega \in \bar{\Omega}_{\kappa'}^{(x,r)}} \tilde{p}_s(\omega) \cdot \mathbf{1}\!\left[C \cap A_{\kappa'}^{(x,r)}(\omega) \neq \emptyset\right] \\
    p_{\text{cfo}}^{(K,x,r,I,z)}(\kappa) &:=
    \begin{aligned}[t]
        \sum_{\omega \in \Omega_\kappa^{(x,r)}} & p_{s}(\omega) \cdot\, \chi^{(x,r,I,z)}(\kappa, \omega) \cdot{} \\
        & \prod_{\kappa' \in K \setminus \{\kappa\}}\!\bigl(1 - p_{\text{ctp}}^{(x,r)}(\kappa', \mathcal{T}^{(x,r,I,z)}(A_\kappa^{(x,r)}(\omega)))\bigr)
    \end{aligned}
    \\
    \mathbb{E}_{\text{fc}}^{(K,x,r,I,z)} &:= \sum_{\kappa \in K} p_{\text{cfo}}^{(K,x,r,I,z)}(\kappa)
\end{aligned}
\end{equation}
}
\end{definition}
$\bar{\Omega}_{\kappa'}^{(x,r)}$ and $\tilde{p}_s$ account for runtime realizations that exceed the plan: before an unselected branch or an exceeded loop bound reveals that case $\kappa'$ is infeasible, $\kappa'$ has already executed every planned activity on the path to that point, and these activities may threaten other cases.
Unselected branches are handled by widening the scenario set: a scenario in $\bar{\Omega}_{\kappa'}^{(x,r)}$ may draw an unselected branch, contributes no activity from the unselected subtree, and continues past the $\bpmnXOR$.
Loop tails are handled by the weight: if a loop requires more than its $r^{\kappa'}_{t(\hat{m})}$ planned iterations, $\kappa'$ still executes all planned iterations, so $\tilde{p}_s$ adds this tail probability ($q^{r}$ for a loop with $r$ planned iterations) to the scenario with the maximum iteration count by omitting its final $(1-q)$ factor.
Since every $\bpmnXOR$ keeps all its branches and every loop tail is absorbed, the weights $\tilde{p}_s$ over $\bar{\Omega}_{\kappa'}^{(x,r)}$ sum to one: every runtime realization of $\kappa'$ maps to exactly one $\omega \in \bar{\Omega}_{\kappa'}^{(x,r)}$ by truncating its loop counts at the plan.
Because the traversal of $\omega$ ignores plan failure, $A_{\kappa'}^{(x,r)}(\omega)$ is a superset of the activities that $\kappa'$ actually executes in this realization; hence $p_{\text{ctp}}$ over-approximates the probability that $\kappa'$ realizes a threat, and $\mathbb{E}_{\text{fc}}$ remains a lower bound.
With $\Omega_{\kappa'}^{(x,r)}$ or $p_s$ instead, $p_{\text{ctp}}$ would understate the threat probability and $\mathbb{E}_{\text{fc}}$ would not be a lower bound: e.g., a case that plans only one of two equally likely branches would threaten the activities overlapping its pre-$\bpmnXOR$ activities with probability $0.5$ although it executes them in every realization.
\subsubsection{The Integrated Multi-Case Scheduling Problem Formulation}
The objective is again to minimize the makespan for a given expected number of feasible cases threshold $E \in [0, \vert K \vert)$.
Unlike the decomposed formulation, the decision variables and constraints are not fixed a priori, as the set of planned activities $A^{(x,r)}$, their precedence relations, and the interval and assignment variables all depend on the selection $(x, r)$.
The formulation is therefore stated in a declarative way over the variable-size sets induced by $(x, r)$; in practice, it can be instantiated by unrolling loops to a sufficient maximum bound with indicator variables disabling unused iterations, or solved directly via constraint-programming solvers that natively support optional and variable-length structures.
Each loop ancestor multiplies the number of XOR and loop occurrences in $T^{(x^{\kappa}, r^{\kappa})}$, so $|\Omega_\kappa^{(x,r)}|$ and $|\bar{\Omega}_\kappa^{(x,r)}|$ grow exponentially in loop-nesting depth.
For each planned activity, we introduce an interval $I_{a},\ a \in A^{(x,r)}$.

{\small
\begin{subequations}\label{eq:cp_integrated}
\begin{align}
\min_{x,r,I,z}\quad  &
\max (\left\{\text{end}(I_a)\mid a \in A^{(x,r)} \right\})
\label{eq:integrated:objective}
\\
\text{s.t.}\quad &
    \mathbb{E}_{\text{fc}}^{(K,x,r,I,z)} \geq E
\\
&
\text{dur}(I_a) = \sum_{\rho \in \mathcal{R}}z_{a,\rho}d_{t(a),\rho}
&& \forall a \in A^{(x,r)}
\label{eq:integrated:dur}
\\
&
\text{start}(I_{a}) \geq \text{end}(I_{a'})
&& \forall a \in A^{(x,r)},\; \forall a' \in pr(a)
\label{eq:integrated:prec}
\\
&
\sum_{\rho \in \mathcal{R}}z_{a,\rho} = 1 && \forall a \in A^{(x,r)}
\label{eq:integrated:resource}
\end{align}
\end{subequations}
}

%% file: sections/04_evaluation.tex
\section{Evaluation}
\label{sec:evaluation}

In this section, we evaluate the performance and the applicability of the proposed approaches.
We first describe the business processes used for evaluation, followed by the implementation details and the evaluation metrics used to compare the approaches. Finally, we present and analyze the results of the evaluation.

\subsection{Business Processes}
We select three business processes with varying characteristics to evaluate the approaches under different conditions. The business processes are selected based on their complexity, resource requirements, and the availability of event logs that contain activity durations.
The selected business processes are:
\textbf{Repair Shop}, an artificial repair center whose process model and branching/looping probabilities are shown in Fig.~\ref{fig:example_schedule};
\textbf{BPIC-14I}, a real-world process from the Business Process Intelligence Challenge (BPIC) 2014, based on IT service management data from Rabobank Group ICT; and
\textbf{BPIC-17W}, a real-world process from BPIC 2017, containing event data from a loan application process at a Dutch financial institute: we selected only the workflow (W) activities, representing the internal work steps performed by bank employees.
The structural characteristics of the business processes are summarized in Tab.~\ref{tab:process_characteristics}.

\begin{table}[h]
\centering
\small
\caption{Structural characteristics of the evaluation business processes.}
\label{tab:process_characteristics}
\vspace{-5pt}
\begin{tabular*}{\textwidth}{@{\extracolsep{\fill}} l cccccccc @{\extracolsep{\fill}}}
\toprule
& $|\bpmnActivity{\text{Act}}|$
& $|\bpmnXOR|$
& $|\bpmnAND|$
& $|\circlearrowright|$
& $|\rightarrow|$
& $|\bpmnActivity{\tau}|$
& Nodes
& Depth \\
\midrule
\textbf{Repair Shop}
& 6 & 1 & 1 & 1 & 2 & 1 & 12 & 5 \\
\textbf{BPIC-14I}
& 6 & 1 & 0 & 0 & 0 & 0 & 7 & 2 \\
\textbf{BPIC-17W}
& 8 & 9 & 3 & 8 & 4 & 17 & 49 & 13 \\
\bottomrule
\end{tabular*}
\vspace{-10pt}
\end{table}

We discovered the process trees for the two real-world business processes using the inductive miner \cite{DBLP:conf/apn/LeemansFA13}, with branching and looping probabilities obtained by replaying the event log on the discovered process tree.
Resource capabilities and processing times are derived from the event logs, where a resource is permitted for an activity if observed executing it at least once, and durations are estimated as medians. To account for varying daily availability, we select a representative day closest to the mean active resource count, yielding 8 of 14 resources for BPIC-14I, and 28 of 145 for BPIC-17W.
For the Repair Shop business process, we generate 9 synthetic resources with varying capabilities.

\subsection{Implementation}
The prototypical implementation\footnote{{\scriptsize \url{https://github.com/ltsstar/BusinessProcessPlanningScheduling/}}} is based on constraint programming models using Google OR-Tools CP-SAT\footnote{{\scriptsize \url{https://developers.google.com/optimization}}}.
Since CP-SAT operates exclusively over integer variables, all floating-point probabilities are scaled to integers with a fixed-point scale factor.

\paragraph{Decomposed Approach}
Since CP-SAT does not natively support power operations, which are required for the activity planning part, the $r$-dependent geometric-series coefficients are precomputed for a sufficient set of $r$ values ($r_{\max}=10$) and implemented as lookup tables.
The total time budget is shared: the planning phase receives at most half of the budget, and the remaining wall-clock time is passed to the scheduling phase.

\paragraph{Integrated Approach}
The integrated formulation is encoded as a single CP-SAT model.
Since CP-SAT does not support a dynamic number of variables, loop unrolling is limited to a maximum iteration bound ($r_{\max}=10$).

\paragraph{Baseline Approach}
A direct comparison with existing approaches is precluded: they either schedule only up to the next uncertain decision point \cite{DBLP:conf/otm/HavurC19} or allocate resources online during execution \cite{DBLP:conf/icpm/KunklerR24,DBLP:journals/is/MiddelhuisBSBAD25}, and thus yield no a-priori schedule until case completion whose makespan and feasibility could be compared against ours.
We therefore compare our approaches with a baseline that selects the single most probable execution using the probabilities of exclusive choices and the expected loop length. Hence, it does not consider the expected feasible cases constraint $E$. The planned activities from the baseline approach are scheduled using the same scheduling logic as for the decomposed approaches, so its scheduling solutions may not be optimal.

\subsection{Evaluation Results}
We ran the evaluation on a machine with an AMD Ryzen 9 PRO 8945HS CPU and 32 GB of RAM, with a 20-minute time limit per run.
The results are summarized in Tab.~\ref{tab:evaluation_results}.
We report the \emph{expected number of feasible cases} of a solution, indicating how closely the solution meets the feasibility constraint $E$, the \emph{makespan} of a solution, the number of \emph{scheduled activities}, reflecting the total resource workload including potentially superfluous activities,
and the solver \emph{duration} and \emph{optimality} flag for each phase.
For those runs marked with ``--'', no feasible solution was found within the time limit.

\input{tables/results.tex}

For the decomposed approach, the planning phase is trivially fast and always optimal, while scheduling is the clear bottleneck. Still, a feasible schedule is found for every configuration except BPIC-17W with $N{=}50$, the most complex process with the highest resource, case, and expected-feasible-case counts.
The integrated approach scales poorly, mostly failing to find a feasible solution for larger configurations within the time limit. Where it does solve (Repair Shop and BPIC-14I with $N{=}5$, BPIC-17W with $N{=}1$), it yields shorter makespans than the decomposed approach, typically at the cost of more scheduled activities.
The baseline approach's feasibility varies by process: for BPIC-14I, with its single exclusive choice, it matches the decomposed approach at the $60\%$ feasibility constraint; for the more complex BPIC-17W, feasibility is extremely low; for Repair Shop, feasibility is lower than the other approaches but with the same makespan as the decomposed approach.

%% file: tables/results.tex
\begin{table}[h]
\vspace{-5pt}
\centering
\small
\caption{{\small Evaluation results {(\scriptsize N : number of cases, E : expected feasible cases constraint, $\mathbb{E}_{\text{fc}}$ : expected feasible cases, MS : makespan (s), SA : scheduled activities, PD : planning duration (s), PO : planning optimality, SD : scheduling duration (s), SO : scheduling optimality, ID : integration duration (s); baseline : single-path selection (argmax branch, expected loop length), independent of E; \textbf{bold}: best value per row among the decomposed and integrated approaches ($\mathbb{E}_{\text{fc}}$ closest to $E$ from above, i.e., meeting the feasibility constraint most tightly, lowest MS, fewest SA); \underline{underlined}: a baseline MS or SA that matches or beats this best, which the single-path baseline may attain while violating the feasibility constraint, i.e., $\mathbb{E}_{\text{fc}}<E$)}}}
\label{tab:evaluation_results}
\vspace{-7pt}
\setlength{\tabcolsep}{3pt}
\resizebox{\textwidth}{!}{%
\begin{tabular}{ccc @{\hspace{3pt}} rrrc @{\hspace{3pt}} rrrrcrc @{\hspace{3pt}} rrrrc}
\toprule
\multicolumn{3}{c}{\textbf{Settings}} & \multicolumn{4}{c}{\textbf{Baseline}} & \multicolumn{7}{c}{\textbf{Decomposed}} & \multicolumn{5}{c}{\textbf{Integrated}} \\
\cmidrule(lr){1-3} \cmidrule(lr){4-7} \cmidrule(lr){8-14} \cmidrule(lr){15-19}
\textbf{P} & \textbf{N} & \textbf{E} & $\mathbb{E}_{\text{fc}}$ & \textbf{MS} & \textbf{SA} & \textbf{SO} & $\mathbb{E}_{\text{fc}}$ & \textbf{MS} & \textbf{SA} & \textbf{PD} & \textbf{PO} & \textbf{SD} & \textbf{SO} & $\mathbb{E}_{\text{fc}}$ & \textbf{MS} & \textbf{SA} & \textbf{ID} & \textbf{Opt} \\
\midrule

\multirow{6}{*}{\rotatebox{90}{\textbf{Repair Shop}}}
                   & 1 & 0.6 & \multirow{2}{*}{0.56} & \multirow{2}{*}{\underline{384}} & \multirow{2}{*}{\underline{5}} & \multirow{2}{*}{\checkmark} & \textbf{0.86} & \textbf{384} & \textbf{6} & 0.01 & \checkmark & 0.01 & \checkmark & \textbf{0.86} & \textbf{384} & \textbf{6} & 0.08 & \checkmark \\
                   & 1 & 0.8 &  &  &  &  & \textbf{0.86} & \textbf{384} & \textbf{6} & 0.01 & \checkmark & 0.01 & \checkmark & \textbf{0.86} & \textbf{384} & \textbf{6} & 0.08 & \checkmark \\
                   & 5 & 3 & \multirow{2}{*}{2.80} & \multirow{2}{*}{960} & \multirow{2}{*}{\underline{25}} & \multirow{2}{*}{--} & 4.30 & 960 & \textbf{30} & 0.01 & \checkmark & 1199 & -- & \textbf{3.01} & \textbf{624} & \textbf{30} & 433 & \checkmark \\
                   & 5 & 4 &  &  &  &  & 4.30 & 960 & \textbf{30} & 0.01 & \checkmark & 1199 & -- & \textbf{4.00} & \textbf{912} & 36 & 1200 & -- \\
                   & 50 & 30 & \multirow{2}{*}{28.00} & \multirow{2}{*}{\underline{9600}} & \multirow{2}{*}{\underline{250}} & \multirow{2}{*}{--} & \textbf{43.00} & \textbf{9600} & \textbf{300} & 0.01 & \checkmark & 1199 & -- & -- & -- & -- & -- & -- \\
                   & 50 & 40 &  &  &  &  & \textbf{43.00} & \textbf{9600} & \textbf{300} & 0.01 & \checkmark & 1199 & -- & -- & -- & -- & -- & -- \\
\midrule

\multirow{6}{*}{\rotatebox{90}{\textbf{BPIC-14I}}}
                   & 1 & 0.6 & \multirow{2}{*}{0.79} & \multirow{2}{*}{\underline{99}} & \multirow{2}{*}{\underline{1}} & \multirow{2}{*}{\checkmark} & \textbf{0.79} & \textbf{99} & \textbf{1} & 0.02 & \checkmark & 0.01 & \checkmark & \textbf{0.79} & \textbf{99} & \textbf{1} & 0.05 & \checkmark \\
                   & 1 & 0.8 &  &  &  &  & \textbf{1.00} & \textbf{175} & \textbf{2} & 0.01 & \checkmark & 0.01 & \checkmark & \textbf{1.00} & \textbf{175} & 3 & 0.05 & \checkmark \\
                   & 5 & 3 & \multirow{2}{*}{3.94} & \multirow{2}{*}{347} & \multirow{2}{*}{\underline{5}} & \multirow{2}{*}{\checkmark} & 3.94 & 347 & \textbf{5} & 0.01 & \checkmark & 0.01 & \checkmark & \textbf{3.04} & \textbf{297} & 18 & 0.26 & \checkmark \\
                   & 5 & 4 &  &  &  &  & 5.00 & 357 & \textbf{10} & 0.01 & \checkmark & 0.03 & \checkmark & \textbf{4.12} & \textbf{310} & 19 & 0.24 & \checkmark \\
                   & 50 & 30 & \multirow{2}{*}{39.35} & \multirow{2}{*}{\underline{1830}} & \multirow{2}{*}{\underline{50}} & \multirow{2}{*}{--} & \textbf{39.35} & \textbf{1830} & \textbf{50} & 0.06 & \checkmark & 1199 & -- & -- & -- & -- & -- & -- \\
                   & 50 & 40 &  &  &  &  & \textbf{49.96} & \textbf{2170} & \textbf{100} & 0.01 & \checkmark & 1199 & -- & -- & -- & -- & -- & -- \\
\midrule

\multirow{6}{*}{\rotatebox{90}{\textbf{BPIC-17W}}}
                   & 1 & 0.6 & \multirow{2}{*}{0.16} & \multirow{2}{*}{\underline{721}} & \multirow{2}{*}{\underline{7}} & \multirow{2}{*}{\checkmark} & 0.61 & 1483 & \textbf{21} & 0.07 & \checkmark & 0.25 & \checkmark & \textbf{0.60} & \textbf{768} & 66 & 35 & \checkmark \\
                   & 1 & 0.8 &  &  &  &  & \textbf{0.80} & \textbf{1895} & \textbf{41} & 0.05 & \checkmark & 1.79 & \checkmark & -- & -- & -- & -- & -- \\
                   & 5 & 3 & \multirow{2}{*}{0.80} & \multirow{2}{*}{\underline{863}} & \multirow{2}{*}{\underline{35}} & \multirow{2}{*}{\checkmark} & \textbf{3.07} & \textbf{1746} & \textbf{105} & 0.06 & \checkmark & 1199 & -- & -- & -- & -- & -- & -- \\
                   & 5 & 4 &  &  &  &  & \textbf{4.02} & \textbf{2258} & \textbf{205} & 0.06 & \checkmark & 1200 & -- & -- & -- & -- & -- & -- \\
                   & 50 & 30 & \multirow{2}{*}{8.03} & \multirow{2}{*}{\underline{5602}} & \multirow{2}{*}{\underline{350}} & \multirow{2}{*}{--} & \textbf{30.67} & -- & -- & 0.07 & \checkmark & -- & -- & -- & -- & -- & -- & -- \\
                   & 50 & 40 &  &  &  &  & \textbf{40.16} & -- & -- & 0.05 & \checkmark & -- & -- & -- & -- & -- & -- & -- \\
\bottomrule
\end{tabular}%
}
\vspace{-10pt}
\end{table}

%% file: sections/05_related_work.tex
\section{Related Work}
\label{sec:related_work}
Planning and scheduling of activities has been addressed in the literature in different domains, including business process management (BPM), operations research (OR), and AI planning.

\paragraph{Scheduling and Resource Allocation in Business Process Management.}
In our prior survey \cite{DBLP:conf/edoc/KunklerSR25}, we identified that existing work on scheduling business process activities often treats the branch taken at exclusive choices as a decision variable and formulates optimization problems that select a single best path (e.g. \cite{van_der_aalst_petri_1996}).
By contrast, in this work, the branch selection is a stochastic outcome necessitating a formulation that plans for sufficiently many paths to guarantee a target level of feasibility.
Some approaches explicitly address the control-flow induced uncertainty by either scheduling as long as certainty is relatively high \cite{DBLP:conf/otm/HavurC19} or presenting online resource-allocation approaches instead of scheduling the whole case \cite{DBLP:conf/icpm/KunklerR24,DBLP:journals/is/MiddelhuisBSBAD25}.

\paragraph{Resource-Constrained Project Scheduling (RCPSP).}
A related prominent problem in OR is the RCPSP, which only involves scheduling a fixed set of activities with known durations and precedence constraints, but does not consider uncertainty in the activity network structure \cite{DBLP:journals/eor/HartmannB10}.
The RCPSP with alternative subgraphs (RCPSP-AS)~\cite{DBLP:journals/ejor/KellenbrinkH15} is similar to our problem formulation: the project network contains alternative execution paths, resembling exclusive choices; however, the choice of alternatives is a decision variable (as in \cite{van_der_aalst_petri_1996}).

\paragraph{AI Planning.}
AI planning approaches have also addressed the problem of planning and scheduling under uncertainty. In particular, conformant probabilistic planning (CPP) seeks solutions that conform to uncertain outcomes, similar to our approach \cite{IntroductionAutomatedPlanning}.
Chance-constrained CPP has also been studied: closest to our approach, \cite{DBLP:conf/aips/SantanaVTWFW16} uses probabilistic simple temporal networks (a graph model without uncertain control flow, but with uncertain activity durations) and LP relaxation to compute a chance-constrained schedule.

%% file: sections/06_conclusion.tex
\section{Discussion and Conclusion}
\label{sec:conclusion}
In this work, we addressed planning and scheduling of business process activities under control-flow induced uncertainty until case completion.
We presented a decomposed approach with sequential planning and scheduling stages, and an integrated approach combining both in a single formulation; both use chance-constrained optimization to select activities such that a predefined case feasibility threshold is met while minimizing the makespan.
The evaluation reveals a trade-off between solution quality and scalability: the integrated approach can yield superior makespans but scales poorly. The decomposed approach, in contrast, reliably finds feasible schedules across nearly all tested configurations.
In practice, scheduling cases until completion rather than only up to the next decision point allows resources to prepare for upcoming activities and enables reliable commitments towards customers.
The feasibility threshold acts as a management lever balancing the risk of rescheduling against reserved but potentially unused capacity, comparable to overbooking.
Since all required inputs can be discovered from event logs, the approaches integrate into existing process-aware information systems with little modeling effort.
A limitation of both formulations is that branch selection for exclusive choices within loops is uniform across iterations, which can lead to suboptimal plans, since later iterations with lower execution probability could benefit from planning fewer branches.
Future work could address iteration-dependent branch selection and incorporate processing time uncertainty alongside control-flow uncertainty.